\documentclass[runningheads,a4paper]{llncs}
\usepackage[mathscr]{euscript}
\usepackage{amssymb}
\usepackage{systeme}
\usepackage{tabularx}
\usepackage{amsmath}
\usepackage{graphicx}
\usepackage{multirow}
\usepackage{tikz}
\usepackage{boldline}
\usepackage{enumerate}
\usepackage{booktabs}
\usepackage{array}
\newtheorem{Remark}{Remark}

\begin{document}

\title{A Protection against the Extraction of Neural Network Models}

\author{Herv\'e Chabanne\inst{1,2}, Vincent Despiegel\inst{1}, Linda Guiga\inst{1,2}}

\institute{Idemia\\
\email{firstname.lastname@idemia.com}
\and
T\'el\'ecom Paris
}
\maketitle

\begin{abstract}
Given oracle access to a Neural Network (NN), it is possible to extract its underlying model. We here introduce a protection by adding parasitic layers which keep the underlying NN's predictions mostly unchanged while complexifying the task of reverse-engineering. Our countermeasure relies on approximating a noisy identity mapping with a Convolutional NN. We explain why the introduction of new parasitic layers complexifies the attacks. We report experiments regarding the performance and the accuracy of the protected NN.
\end{abstract}

\section{Introduction}
Accurate Neural Networks require a carefully selected architecture and a long training on a large database. Thus, NN models' architecture and parameters are often considered intellectual property. Moreover, the knowledge of both the architecture and the parameters make adversarial attacks -- among other kinds of attacks -- easier: an attacker can easily generate small input noise that is undetectable by the human eye but still changes the model's predictions \cite{adv,adv_survey}.

Several papers \cite{DBLP:journals/corr/abs-2003-04884,DBLP:conf/fat/MilliSDH19,DBLP:journals/corr/abs-1910-00744,DBLP:journals/corr/abs-1909-01838} have exploited the fact that the layers of a ReLU Neural Network (NN) are piecewise linear functions to extract its underlying model's weights and architecture. 
Indeed, hyperplanes -- separating the spaces where the ReLU NN is linear -- split the model's input space, and recovering the boundaries formed by the hyperplanes enables the extraction of its weights and architecture. 

Here, we show how to modify the naturally induced division of the input space by inserting parasitic layers between the NN layers. Our parasitic layers are going to approximate a function close to the identity mapping, following \cite{pni}. Since this adds new polytopes -- whose boundaries are the various hyperplanes --, it leaves the flow of data within the victim NN mostly unchanged, while disrupting the geometry accessible for extraction. 

After finishing this introduction, we recall the aforementioned extraction of RELU NN in Sec.~\ref{attack}. We show in Sec.~\ref{identity}, following \cite{DBLP:journals/corr/abs-1902-04698}, how to approximate the identity through a Convolutional NN (CNN). We then describe our protection proposal in Sec.~\ref{proposal}. In Sec. \ref{security}, we explain how adding a CNN approximating a noisy identity mapping mitigates model extraction attacks on NNs.  
Sec.~\ref{experiments} reports our experiments regarding the deterioration of performances and accuracy due to the addition of parasitic layers 

\subsection{Background}
Today, Neural Networks (NNs) are used to perform all kinds of tasks, ranging from image processing \cite{vgg} to malware detection \cite{malware}. Neural Networks are algorithms that, given an input $x$, compute an output $o$ usually corresponding to either a classification or a probability. NNs are organized in layers. Each layer contains a set of neurons. Neurons of a given layer are computed based on a subset from the previous layer's parameters and parameters called weights. 

There are different types of layers. Among those are:
\begin{itemize}
	\item Fully connected layers: Each neuron from a layer $l_i$ is connected to all neurons from layer $l_{i+1}$. Thus, a neuron $\eta_k^i$ in a layer $l_i$ is computed as follows: $\eta_i = \sum_{j=1}^n \eta_j^{i-1} w_j^i$ where $\{\eta_j^{i-1}\}_j$ are the $n$ neurons from the previous layer and $\{w_j\}_j$ are the layer's weights. 
	\item Convolutional layers: These layers compute a convolution between one -- or several -- filter F and windows from the input, as follows:
	$$
	O_{i, j} = \sum_{k=1}^{h}\sum_{l=1}^{w} X_{i+k, j+l}\cdot F_{k, l}
	$$
	
	The elements of the filter are the weights of the layer. The number of filters is the number of output channels. An input can have several channels. For instance, in image processing, the input of a model is usually an image with three channels, corresponding to the RGB colors.
	\item Batch Normalization layers: These layers aim at normalizing the input. To achieve this, they learn the mean  and standard deviation over mini-batches of input, as well as $\gamma$ and $\beta$ parameters, and return:
	$$
	O_i = \gamma_i \times \frac{x_i - E_B}{\sqrt{V_B + \epsilon}} + \beta_i
	$$ 
	where $x = (x_1, ..., x_n)$ is the layer's input and $E_B$ and $V_B$ are the learnt mean and variance respectively.
	These layers aim at removing the scaling factor introduced through the previous layers. They make training faster and more efficient. 
\end{itemize}

While the various layers of an NN are linear, each layer is followed by an activation function, applied to all of the layer's neurons. The activation function is used to activate or, on the contrary, deactivate some neurons. One of the most popular and simplest activation function is ReLU, defined as the maximum between 0 and the neuron.

NNs only composed of fully connected layers are called Fully Connected Networks (FCNs), while those which are mainly composed of convolutional layers are called Convolutional Neural Networks (CNNs).

A ReLU NN is a NN constituted by linear layers followed by ReLU activation functions.

Let us note that another common layer type is the pooling layer, whose goal is to reduce the dimensionality. Since the attacks at hand do not take those layers into account, we also put ourselves in the context where pooling layers are not considered. 

\subsection{Related Works}
Different kinds of reverse engineering approaches have been introduced. Batina et al. recover NNs' structure through side channels, i.e. by measuring leakages like power consumption, electromagnetic radiation, and reaction time \cite{DBLP:conf/uss/BatinaBJP19}. These measurement attacks are common for embedded devices (e.g. smartcards). Fault attacks, which are also a typical threat to smartcards, are transposed to find NN models in \cite{DBLP:journals/corr/abs-2002-11021}. A weaker approach where the victim NN shares its cache memory with the attacker in the cloud is taken in \cite{DBLP:journals/corr/abs-2002-06776,DBLP:journals/corr/abs-1808-04761}. The protections to thwart these attacks are related to the victim NN implementation. As we here consider oracle access attacks, our countermeasures have to modify the NN's architecture itself.

A more detailed explanation of the attacks \cite{DBLP:journals/corr/abs-2003-04884,DBLP:conf/fat/MilliSDH19,DBLP:journals/corr/abs-1910-00744,DBLP:journals/corr/abs-1909-01838} is given in the next Section. 

It should be noted that the abstract model of NNs that we are looking at here has been introduced by \cite{DBLP:journals/corr/abs-1901-10861} while in the different context of adversarial examples. Similarly to \cite{DBLP:journals/corr/abs-1901-10861}, the authors of \cite{deepfool} use the hyperplanes introduced by the activation functions and the class boundaries they form in order to accurately compute adversarial examples, as well as the robustness of the original model. While this is not the primary application of our idea, its transposition to thwart adversarial examples seems intriguing. As a matter of fact, we are going to gauge the efficiency of our countermeasure thanks to adversarial attacks. 

\section{Extraction of Neural Network Models} \label{attack}
Several attacks \cite{DBLP:journals/corr/abs-2003-04884,DBLP:conf/fat/MilliSDH19,DBLP:journals/corr/abs-1910-00744,DBLP:journals/corr/abs-1909-01838} have managed to recover a ReLU NN's weights. These attacks rely on the fact that ReLU is piecewise linear. 

The attack model in \cite{DBLP:conf/fat/MilliSDH19}, \cite{DBLP:journals/corr/abs-1909-01838} and \cite{DBLP:journals/corr/abs-1910-00744} is as follows:
\begin{itemize}
	\item The victim model corresponds to a piecewise linear function
	\item The attacker can query the model
	\item The attacker aims at recovering the weights (and, in some cases \cite{DBLP:journals/corr/abs-1910-00744}, the architecture) of the victim model
	\item The victim model is composed of linear layers (such as FC ones), as well as ReLU activation functions.
\end{itemize}
Furthermore, \cite{DBLP:journals/corr/abs-1910-00744} also assumes that the attacker does not know the structure (i.e. the number of neurons per layer) of the victim NN. 
In the case of \cite{DBLP:journals/corr/abs-2003-04884}, the authors assumed that the attacker had access to the architecture, but not the weights.  However, the authors mention their belief that the piecewise linearity of the NN is the only assumption fundamental to their work, even though they do not prove it in their paper. 

This attack model corresponds to the case of online services, for instance, where users can query a model and get the output, but they do not have access to the architecture and parameters of the model. 

\cite{DBLP:journals/corr/abs-2003-04884} is the only paper so far that proves the practicability of its attack for more than 2  layers of a given neural network, even though the theory of \cite{DBLP:conf/fat/MilliSDH19} applies to arbitrarily deep neural networks. Moreover, \cite{DBLP:journals/corr/abs-2003-04884} provides a much higher accuracy with much fewer queries to the victim we want to protect. 

Let $\mathcal{V}(\eta, x)$ denote the input of neuron $\eta$, before applying the ReLU activation function, when the model's input is $x$. For a given neuron $\eta$ at layer $l$, let us define its critical point as follows:
\begin{definition}
When, for an input $x$, $\mathcal{V}(\eta, x) = 0$, the neuron $\eta$ is said to be at a critical point. Moreover, $x$ is called a witness of $\eta$ being at a critical point. 
\end{definition}
Finding at least one witness for a neuron $\eta$ enables the attacker to compute $\eta$'s critical hyperplane.
\begin{definition}
A bent critical hyperplane for a neuron $\eta$ is the piecewise linear boundary $\mathcal{B}$ such that $\mathcal{V}(\eta, x) = 0$ for all $x \in \mathcal{B}$.  
\end{definition}

All three attacks recover the weights of each layer thanks to the following steps:
\begin{enumerate}
	\item Identify critical points and deduce the critical hyperplanes
	\item Filter out critical points from later layers
	\item Deduce the weights up to the sign and up to an isomorphism
	\item Find the weight signs
\end{enumerate}
Although the way critical points are found and filtered out differ from an article to the other, all methods use the piecewise linearity of the ReLU activation. The main element in those attacks resides in the fact that each neuron is associated to one bent critical hyperplane (that exists because of the ReLU activation function), corresponding to the neuron's change of sign. That hyperplane's equation is what enables the attacker to deduce the weights. 

Let us detail the attack in \cite{DBLP:journals/corr/abs-2003-04884}, as it is the most accurate and requires the fewest queries to the victim model so far. 


\subsubsection{Finding critical points}
\begin{figure}
\centering
\begin{tikzpicture}
	\draw (-2.25, 4.2) node{$\eta_0$};
	\draw[thick] (-2, 4) -- (0, 3) -- (1, 2.5) ;
	\draw (0.75, 4.75) node{$\eta_1$};
	\draw[thick] (0.75, 4.5) -- (0, 3) -- (-1, 1) ;
	\draw (-1/3+0.2, 0.75) node{$\eta_2$};
	\draw[thick] (-1/3, 1) -- (-1, 5);
	\draw (-2.2, 2) node{$l$};
	\draw[dashed, |-|, red] (-2, 2.2) -- (-1, 2.7) -- (0, 3.2) -- (1, 3.7); 
	\draw (2/15, 49/15) node{\textbf{\textbackslash}};
	\draw (-42/65, 187/65) node{\textbf{/}};
	\draw (-0.2, 31/10) node{\textbf{\textbackslash}};
	
\end{tikzpicture}
\caption{Hyperplanes for three neurons in the first layer. The dashed red line $l$ enables the attacker to find the critical points indicated by the slashes.}
\label{fig:first_layer_hyperplanes}
\end{figure}
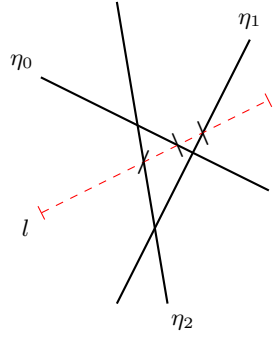

The attacker chooses a random line $\mathit{l}$ from the input space. Looking for non linearities through binary search in a large interval in that line enables the attacker to find several critical points (see Fig.~\ref{fig:first_layer_hyperplanes}). 

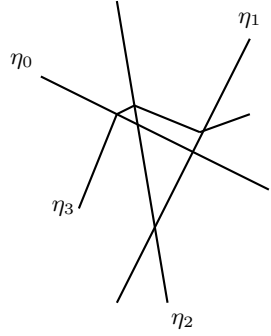
\begin{figure}
	\centering
	\begin{tikzpicture}
		\draw (-2.25, 4.2) node{$\eta_0$};
		\draw[thick] (-2, 4) -- (0, 3) -- (1, 2.5) ;
		\draw (0.75, 4.75) node{$\eta_1$};
		\draw[thick] (0.75, 4.5) -- (0, 3) -- (-1, 1) ;
		\draw (-1/3+0.2, 0.75) node{$\eta_2$};
		\draw[thick] (-1/3, 1) -- (-1, 5);
		\draw (-1.7, 9/4) node{$\eta_3$};
		\draw[thick] (-1.5, 9/4) -- (-1, 3.5) ;
		\draw[thick] (-1, 3.5) -- (-10/13, 47/13);
		\draw[thick] (-10/13, 47/13) -- (6/65, 212/65);
		\draw[thick] (6/65, 212/65) -- (0.75, 3.5);
	\end{tikzpicture}
	\caption{Hyperplanes are bent by boundaries from previous layers. For instance, $\eta_3$'s hyperplane on the second layer is bent by the hyperplanes of $\eta_0$, $\eta_1$ and $\eta_2$ on the first layer.}
	\label{fig:later_layers_hyperplanes}
\end{figure}

However, the attacker knows neither what neurons these critical points are witnesses for, nor the said neurons' layer. Neurons from the first layer yield unbent hyperplanes, while those in the following layers are bent by the several previous ReLUs (see Fig.~\ref{fig:later_layers_hyperplanes}). 

\subsubsection{Recovering the weights up to a sign}
As seen before, the attacker has a set of witnesses for neurons in all layers. She can then carry out a differential attack in order to recover the weights and biases up to a sign. 

Let us describe the attack on a simple case where the model only has one hidden layer, and the input vector space is $\chi = \mathbb{R}^N$. Let $x^*$ be a witness for neuron $\eta^*$ being at a critical point. Define $\{e_i\}$ as the set of standard basis vectors of $\chi$. The attacker computes:
$$
\alpha_+^i = \frac{\partial f(x)}{\partial e_i} \Bigg |_{x=x^* + e_i} \text{ and } \alpha_-^i = \frac{\partial f(x)}{\partial e_i} \Bigg |_{x=x^* - e_i}
$$
Then, because the activation function is $ReLU(x) = max(0, x)$, we have that: $\alpha_+^i - \alpha_-^i = \pm A_{j, i}^{(1)} \cdot A^{(2]}$.
Thus, by computing:
$$
\frac{\alpha_+^i - \alpha_+^i}{\alpha_+^1 - \alpha_+^1}
$$
for all $i$, the attacker gets the weights up to a multiplicative scalar.

In the general case where the NN is deeper, and for a layer $j$, the attacker computes second partial derivatives $y_i = \{\frac{\partial^2 f}{\partial \delta_i^2}\}$ instead of the simple ones, where the $\delta_i$ take random values. She then solves a system of equations: $h_i \cdot w = y_i$, where $h_i$ is the value of the previous layer -- after the $ReLU$ -- for an input model $x^* + \delta_i$. Let us note that the attacker does not know whether neuron $\eta^*$ is in the current layer. She therefore solves the system of equations for all layers, and only keeps the solution that appears most often. 
The biases can then easily be deduced from the weights.

To differentiate critical points of the current layer from other critical points, the differential attack is carried out on all the critical points and the attacker filters out the wrong critical points by observing the resulting traces. 

\subsubsection{Recovering weight signs}
In this step, the attacker proceeds recursively. The attacker has a set $\mathcal{S}$ of witnesses for unknown neurons (as found in the previous step). 

Let us suppose the attacker has managed to recover the correct model up to layer $j-1$, as well as the weights up to sign for layer $j$. 
Let us define the polytope at layer $j$ containing $x$ as:
$$
\mathcal{S} = \{x + \delta \text{ s.t. } sign(\mathcal{V}(\eta, x)) = sign(\mathcal{V}(\eta, x+\delta))\}
$$
Thus, this polytope corresponds to the open, convex subspace shaped by the critical hyperspaces. 

The attacker can easily filter out the critical points $x$ from previous layers since she already recovered the weights and biases up to layer $j$.

To filter out witnesses from layers deeper than $j+1$, the attacker relies on the fact that the polytopes of two distinct layers have a different shape with high probability. 

Finally, the attacker recovers the sign of the weights through brute force using layer $j+1$'s witnesses. 
Let us note that when the victim NN is contractive, the sign recovery can be less expensive. 

Thus, the attacker can recover the victim model's parameters recursively over the depth of the considered layer as described in the previous paragraphs. Moreover, even though the number of queries is linear, the work required is exponential, as explained in the previous paragraph.

\section{Approximating the Identity thanks to CNNs} \label{identity}
Our proposal is based on adding parasitic layers to the model we want to protect, and for those layers, we rely on a CNN approximating the identity. It results in the addition of dummy hyperplanes, as explained in Sec.~\ref{security}. However, it is not enough to thwart the attack at hand. In order to mitigate the said attack, our parasitic CNNs approximate the identity to which we add a centered Gaussian noise. Sec.~\ref{security} details how this additional noise ensures that the introduced hyperplanes lie in the same space as the original ones. 

Since CNNs are intrinsically nonlinear, approximating the identity -- the simplest linear mathematical function -- would appear to be a difficult learning task. However, thanks to the bias and the piecewise linearity of ReLU, CNNs manage to avoid the obstacle of the hyperplanes by shifting the input to a space where the activation function is linear. Therefore, CNNs manage to approximate the identity very accurately. 

The simplicity of the task at hand is demonstrated in \cite{DBLP:journals/corr/abs-1902-04698}. Indeed, the authors of \cite{DBLP:journals/corr/abs-1902-04698} manage to approximate the identity mapping using CNNs with few layers, few channels and only one training example from the MNIST dataset \cite{mnist}. 

First, they observe that while both CNNs and FCNs could approximate the identity on digits well when trained on three training examples from the MNIST dataset \cite{mnist}, only CNNs generalize to examples outside of the digits scope. Moreover, they state that this bias can still be observed when the models are trained with the whole MINST training set. 

In order to better characterize the observed bias, the authors take the worst case scenario: they only train FCNs and CNNs on a single training example. Contrary to what they expected, architectures that are not too deep manage some kind of generalization: FCNs output noisy images for inputs that are not the training example, while CNNs still manage to approximate the identity. Moreover, FCNs tend to correlate more to the constant function than to the identity. The output of CNNs' correlation with the identity function decreases with a smaller input size and a higher filter size.

The authors of \cite{DBLP:journals/corr/abs-1902-04698} show -- by providing possible filter values -- that in their case, if the input has $n$ channels, $2n$ channels suffice to approximate the identity mapping with only one training example. They also note that adding output featuremaps does help with training. Moreover, they use $5 \times 5$ filters for all their CNNs' layers. Finally, they explain that even though 20-layer CNNs can learn the identity mapping given enough training examples, shallower networks learn the task faster and provide a better approximation. 

This ability of CNNs to learn the identity mapping from only one training example from the MNIST dataset and to generalize it to other datasets shows the simplicity of the task. We explain in Sec. \ref{proposal} and \ref{security} how this fact impairs the defense when the parasitic CNN approximates the identity mapping, and the necessity to approximate a noisy identity as well as to apply some constraints on the CNN's parameters. 

\section{Our proposal} \label{proposal}
Let us consider a victim ReLU NN. The attack scenario described in Sec.~\ref{attack} is based on the bent critical hyperplanes induced by the ReLU functions in the model. In \cite{DBLP:journals/corr/abs-2003-04884}, the bent hyperplanes are especially used in the case of expansive NNs -- i.e. for which a preimage does not always exist for a given value in the output space --, in order to filter out witnesses that are not useful to the studied layer. In order to make the attacker's task more complex, we propose to add artificial critical hyperplanes. Adding artificial hyperplanes would make the attack more complex: the attacker would have to filter out the artificial hyperplanes as well as the other layers' hyperplanes. 

As explained in Sec.~\ref{identity}, CNNs can provide a very good approximation of the identity mapping. Moreover, they generalize well: with only a single trainable example from the MNIST dataset, CNNs up to 5 layers deep can still reach the target. 

We propose to add dummy hyperplanes through the insertion, between two layers of the model to protect, of parasitic CNNs approximating an identity where a centered Gaussian noise has been added. The CNNs we add select $nb$ neurons at random from the output of the previous layer, and approximate a noisy identity, where $nb$ is smaller or equal to the output size of the previous layer.  

Since CNNs approximate the identity well, inserting CNNs approximating the identity yields hyperplanes that do not impact a potential attacker. Indeed, as will be further detailed in Sec. \ref{security} the CNN can make sure that the introduced CNNs are either far from the original ones or, on the contrary, very close and almost parallel to the original ones. In these cases, with high probability, an attacker would not notice the added layers, and would therefore be able to easily carry out her attack. Therefore, we need to apply further constraints on the parasitic CNNs. Instead of CNNs approximating the identity, we propose to insert CNNs approximating the identity where a centered Gaussian noise is added. Furthermore, these CNNs are trained with constraints on some of their parameters. Sec. \ref{security} explains why the addition of the noise helps make the injected hyperplanes noticeable by a potential attacker. 

\begin{Remark} Note that we can think of a dynamic addition of parasitic CNNs approximating a noisy identity mapping. For instance, considering a client-server architecture where the server is making predictions; from a client query to another, different parasitic CNNs  can be added in random places of the server's NN architecture, replacing the previous ones.
\end{Remark}

Furthermore, the small CNN we add does not act on all neurons. This yields two advantages:
\begin{itemize}
	\item The added CNN considered can be small, implying fewer computations during inference
	\item We can add different CNNs to different parts of the input, to further increase the difference in behavior between neurons
\end{itemize}
Fig.~\ref{fig:idNN} shows an example of adding such an identity CNN between the first and the second layer of an NN with only one hidden layer.

\def\layersep{2.5cm}
\def\seclayersep{2.5cm}
\def\thirdlayersep{9cm}
\begin{figure}[htb]
\centering
\resizebox{\textwidth}{!}{%
\begin{tikzpicture}[shorten >=1pt,->,draw=black!50, node distance=\layersep]
    \tikzstyle{every pin edge}=[<-,shorten <=1pt]
    \tikzstyle{neuron}=[black, circle, minimum size=17pt,inner sep=0pt]
    \tikzstyle{input neuron}=[neuron];
    \tikzstyle{output neuron}=[neuron];
    \tikzstyle{hidden neuron}=[neuron];
    \tikzstyle{annot} = [text width=4em, text centered]

    \foreach \name / \y in {1,...,8}{
        \node[input neuron, pin=left:Input \#\y] (I-\name) at (0,-\y) {};
    	\draw[input neuron, pin=left:Input \#\y] (0, -\y) circle (8pt) {};}
		
    \foreach \name / \y / \v in {1/3/2.5,2/4/3.5, 3/5/4.5, 4/6/5.5, 5/7/6.5}{
        \path[yshift=0.5cm]
            node[hidden neuron] (H-\name) at (\thirdlayersep,-\y cm) {};
        \draw[hidden neuron] (\thirdlayersep, -\v cm) circle (8pt) {};
            }

	\draw[black] (0.5, -0.8) rectangle (2, -4.1) node[midway, align=center, text width=1cm] {Identity CNN};
	
    \node[output neuron,pin={[pin edge={->}]right:Output}, right of=H-3] (O) {};
	\draw[output neuron,pin={[pin edge={->}]right:Output}, right of=H-3] (\thirdlayersep, -4.5) circle (8pt) {};
	
	
	\foreach \source in {1,...,4}{
		\path (I-\source) edge (0.5, -\source);
	}
	
	 \foreach \name / \y / \v in {1/1/1,2/2/2, 3/3/3, 4/4/4}{
        \node[hidden neuron] (N-\name) at (\seclayersep,-\y) {};
    	\draw[hidden neuron] (\seclayersep, -\v) circle (8pt) {};}
        
    \foreach \source in {1,...,4}
    	\path (2, -\source) edge (N-\source);
	
	\foreach \source in {1,...,4}
		\foreach \dest in {1,...,5}
			\path (N-\source) edge (H-\dest);	
		
	\foreach \source in {5,...,8}
		\foreach \dest in {1,...,5}
			\path (I-\source) edge (H-\dest);

    \foreach \source in {1,...,5}
        \path (H-\source) edge (O);

    \node[annot,above of=H-1, node distance=1cm] {Hidden layer};
\end{tikzpicture}
}%
\caption{Neural Network with one hidden layer where a CNN approximating the identity has been added to approximate the first fours input neurons}
\label{fig:idNN}
\end{figure}
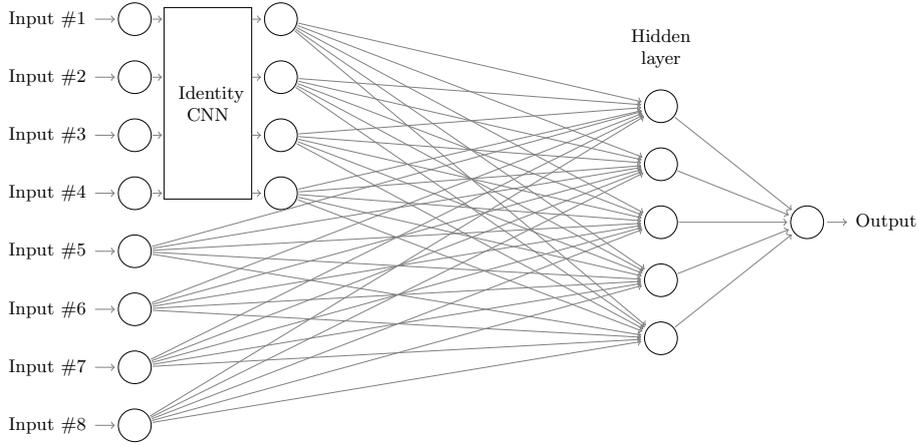

\begin{figure}[ht]
	\centering
	\includegraphics[scale=0.45]{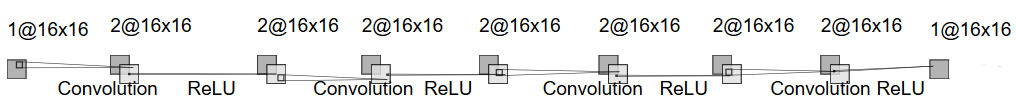}
	\caption{Parasitic CNN with 4 convolutional layers, with a ReLU activation function after each convolution.}
	\label{CNN_no_batch}
\end{figure}

\begin{figure}[ht]
	\centering
	\includegraphics[scale=0.39]{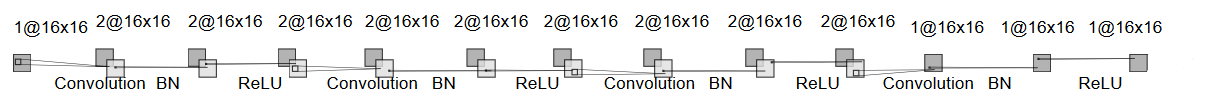}
	\caption{Parasitic CNN with 4 convolutional layers, with a batch normalization layer (BN) and a ReLU activation function after each convolution}
	\label{CNN_batch}
\end{figure}

The CNN we add in our experiments consists of four hidden convolutional layers, with $5 \times 5$ filters (see Fig. \ref{CNN_no_batch}). In some cases, we add a batch normalization layer after each covolutional one (see Fig. \ref{CNN_batch}). Indeed, as recalled in Sec.~\ref{identity}, a CNN with few layers and $5 \times 5$ filters can already approximate the identity on $28 \times 28$ inputs with a single training example. Thus, such a CNN is well adapted to learning the identity mapping on $nb$ neurons, where $nb$ is smaller or equal to the size of the previous layer's output. 
When the CNN receives the set of neurons from the considered layer, it first reshapes it into a square input with one channel, so that it is adapted to convolutional layers. 

Moreover, for the much harder task tackled by the authors of \cite{DBLP:journals/corr/abs-1902-04698}, for an input with $n$ channels, $2n$ channels in the intermediary layers are enough to get a good approximation of the identity, even though more channels improve the accuracy. Since we do not constrain ourselves to training our CNN with a single example, we can limit the number of channels in the hidden layers to two -- because we consider inputs with one channel. This enables us to minimize the number of additional computations for the dummy layers, with only a slight drop in the original model's accuracy.

\section{Complexity of Extraction in the Presence of Parasitic Layers} \label{security}
Adding a convolutional layer with $k$ layers as described in the previous section results in adding $k$ layers to the architecture while keeping almost the same accuracy. If those $k$ layers add critical hyperplanes, then the complexity of extraction increases. 

In this section, we first consider a CNN approximating the identity mapping added after the first layer in the victim NN. We further assume that there are fewer neurons in the second layer than in the first. We prove that in that case, the identity CNN does add hyperplanes with high probability. Then, we explain the need to approximate the need to approximate a noisy identity mapping rather than the identity itself. 

Let us suppose we add a CNN Identity layer that takes $n \times n$ inputs, and the original input size is $m$. Let $\{F_{i, j}\}_{1 \leq i \leq k \text{, } 1 \leq j \leq k}$ be its associated filter. This would result in the following weight matrix C:

\[
\systeme*{C_{i \times n + j, (i+l) \times n + j + h} &= F_{l, h} \text{ } \forall 1 \leq i \text{, } j \leq n-k+1 \text{ and } 1 \leq l \text{, } h \leq k, C_{i, i} &= 1 \text{ } \forall i \geq (n-k+1) \times (n-k+1), C_{i, j} &= 0 \text{ otherwise }}
\]
Here, without loss of generality, we consider there is no padding. 

This new layer adds at most $n \times n$ bent hyperplanes. This number decreases if two neurons $\eta_i$ and $\eta_j$ share the same hyperplane. 

Let $\mathcal{V}(\eta_i, x)$ be the value of $\eta_i$ before the activation function, if the model's input is $x$.

We need to consider two cases:
\begin{enumerate}
	\item $\eta_i$ and $\eta_j$ are in different layers. Let us suppose that $\eta_i$'s layer is $l$ and that $\eta_j$'s layer is $l+1$. If the layers are not consecutive, the $\eta_j$'s hyperplane is bent by $ReLUs$ from the layers in between, making the probability of the two hyperplanes matching very low. 
	\item $\eta_i$ and $\eta_j$ are in the same layer
\end{enumerate}

\subsection{First case: $\eta_i$ is on layer $l$ and $\eta_j$ is on layer $l+1$}

Let us suppose that $\eta_i$ is on the first layer, and $\eta_j$ is on the second one. The output $z(x)$ of the first layer, for $x \in \chi$ is:
$$
z(x) = A^{(1)}x + \beta^{(1)}
$$
In this proof, the rows of $A^{(1)}$ are supposed to be linearly independent. This is an assumption made in \cite{DBLP:conf/fat/MilliSDH19} and in \cite{DBLP:journals/corr/abs-1909-01838}. As stated in \cite{DBLP:journals/corr/abs-1909-01838}, this is likely to be the case when the input's dimension is much larger than the first layer's. The authors of \cite{DBLP:journals/corr/abs-2003-04884} state that it is the case in most ReLU NNs, but not necessarily the most interesting ones. However, the general attack in \cite{DBLP:journals/corr/abs-2003-04884} for the cases where the model to protect is not contractive is more complex, and requires a layer by layer brute force attack for the sign recovery.

The output of the second layer is:
$$
Out = C \cdot ReLU(z(x)) + \beta^{(2)}
$$
Since the rows of $A^{(1)}$ are supposed to be linearly independent, for a given vector V, there exists a solution $x^*$ such that $z(x^*) = V$, by the Rouch\'e-Capelli theorem. If we select $V$ so that $V_i \geq 0 \text{ } \forall i \leq m$, then $V$ is not affected by the $ReLU$. We can therefore select a vector $V$ such that, letting  $k$ be the convolutional layer's filter size:
\begin{align*}
&V_{(\lfloor \frac{j}{n} \rfloor+h) \times n + j\%n + l} = 0 \text{ } \forall 1 \leq l, h \leq k \text{ (where $j \% n$ means $j$ modulo $n$)}\\
&\text{ except for one value } i' \text{ that is not } \eta_i, \text{ where } V_{i'} = 1 
\end{align*}
 
Since this second layer is a convolutional one, $\beta_i^{(2)}$ is the same for all $i$ on a given channel, denoted $\beta$.
The window considered to compute $\eta_j$ is zeroed out, except for one value. The filter weight associated to that value needs to be $-\beta$ to nullify $\eta_j$. Since we can repeat the process for all values of the window that are not $\eta_i$, all the filter weights except for that associated to $\eta_i$ need to be $-\beta$ except for the one associated with $\eta_i$. This is not the case with high probability. Thus, with high probability, $\eta_i = 0$ does not imply that $\eta_j = 0$. 

For deeper layers, even though we cannot select any vector $V$, it is highly unlikely for the following equation to happen:
$$
z_i(x) = 0 \iff C_j ReLU(z(x) + \beta^{(2)}) = 0
$$ 
When $\eta_i$ is not in the window used to compute $\eta_j$, it is even less likely to be the case. 

Therefore, two neurons on different layers are very likely to have different critical hyperplanes. 



\subsection{Second case: $\eta_i$ and $\eta_j$ are in the same layer}

Let us suppose that $\eta_i$ and $\eta_j$ are in layer $l$. Let $l$ be the first convolutional layer. Moreover, let us suppose that the CNN is set after the first layer of the model we want to protect. Then $l$'s input is:
$$
z(x) = ReLU(A^{(1)}x + \beta^{(1)})
$$
where $x$ is the model's input. 

Let us also suppose, without loss of generality, that $j > i$. This means that the windows used to compute the two neurons are not identical. With high probability, one of the filter values associated with the disjoint window values is nonzero. For simplicity, and without loss of generality, let us suppose, in what follows, that $F_{1, 1}$ is such a filter value. Thus, in what follows, we suppose that $F_{1, 1} \neq 0$. 
\paragraph{Case where $\beta$ = 0 :} As explained before, we can find $x^*$ such that $z(x^*)_{\lfloor \frac{i}{n} \rfloor \times n + i \% n} = 1$ and $z(x^*)_h = 0$ otherwise. Since $j > i$, $z(x^*)_{\lfloor \frac{i}{n} \rfloor \times n + i \% n}$ is not in the window used to compute $\eta_j$, but it is in $\eta_i$'s window. In this case, $\eta_i \neq 0$ and $\eta_j = 0$. Thus, $\eta_i$ and $\eta_j$ do not share the same critical hyperplane. 

\paragraph{Case where $\beta \neq 0$ :} If $\beta \neq 0$ , we cannot directly apply the previous reasoning. Let $x^*$ be a witness for $\eta_j$ being at a critical point. Let us show that we can find an input $x^{**}$ such that $\eta_j = 0$ but $\eta_i \neq 0$. 

If $x^*$ already satisfies this property, our work is done. Otherwise, $x^*$ is such that $\eta_i = \eta_j = 0$. 
As explained before, there exists an input to the NN $x'$ such that $(A^{(1)} \cdot x')_{\lfloor \frac{i}{n} \rfloor \times n + i \% n} = a$ with $a > 0$ and $(A^{(1)} \cdot x')_h = 0$ otherwise. Then, by piecewise linearity of $z$, we have, for $a$ large enough, that $z(x^* + x')_{\lfloor \frac{i}{n} \rfloor \times n + i \% n} > z(x^*)_{\lfloor \frac{i}{n} \rfloor \times n + i \% n}$. Moreover, for all other indices $h$, $z(x^* + x')_h = z(x^*)_h$. Let us consider $x^{**} = x^* + x'$. We have that $z_{\lfloor \frac{i}{n} \rfloor \times n + i \% n}$ is not in $\eta_j$'s window, which means that $\eta_j$ remains unchanged and $\eta_j = 0$ when the NN's input is $x^{**}$. On the other hand, $\eta_i$'s value changes since one of its window values changes and $F_{1, 1} \neq 0$. Thus, $\eta_i \neq 0$. 
Therefore, we can indeed find $x^{**}$ such that $\eta_j = 0$ but $\eta_i \neq 0$.
 
\paragraph{ } Let us now consider the case where $\eta_i$ and $\eta_j$ are on deeper layers, in which case the previous proof does not hold. Let $i = i_1 \times n + i_2$ and $j = j_1 \times n + j_2$, where $i_1 \neq j_1$, or $i_2 \neq j_2$, or both. Let also $F$ be the filter of the considered convolutional layer, of size $k \times k$.

If $\eta_i$ and $\eta_j$ share the same hyperplane, then whenever $z$ is such that $C_i z + \beta = 0$, we have that:
\begin{equation} \label{eq}
\sum_{l = 1}^{k} \sum_{l = 1}^{k} F_{l, h} \left( z_{(i_1 + l) \times n + i_2 + h} - z_{(j_1 + l) \times n + j_2 + h} \right) = 0
\end{equation}
Since Eq. \ref{eq} needs to hold for all the $z$ that are on the hyperplane, this equation is very unlikely to hold.

Therefore, with a very high probability, no two neurons in the same layer share the same hyperplane. 




\subsection{Approximating a Gaussian noise}
As explained before, adding CNNs approximating the identity to a victim neural network adds hyperplanes. However, this does not necessarily lead to an increased complexity for the extraction attacks at hand. Indeed, the identity CNN might avoid the complexity of the task by isolating the newly introduced hyperplanes -- meaning the critical points are far from the input space --, or very close and parallel to the original hyperplanes -- i.e. the critical points correspond to a small translation from the original points. The first case can be achieved by increasing the bias in the convolutional layers, so that all values are made significantly positive. This ensures that no value is zeroed out during the computations. The last layer's bias then translates the values back to their original position. In both cases, the attacker would not notice the introduced hyperplanes, thus defeating the purpose of the parasitic CNN.

The authors of \cite{pni} inject normal noise during a model's training as a way of mitigating adversarial attacks. They introduce a parameter, $\alpha$, trained along the original model so that $\alpha \times \mathcal{N}$ -- where $\mathcal{N}$ is a fixed Gaussian noise -- is added to some layers. Furthermore, they add adversarial examples to the training set to prevent $\alpha$ from converging to 0. 

Similarly to \cite{pni}, we inject noise into our layers in order to avoid cases where the CNN we add is not detectable by an attacker. However, our method separates the training of the added CNN from that of the model to protect. Having to train the original model for each parasitic CNN would result in too much overhead. We inject a fixed Gaussian noise to the labels during the training of our CNN approximating the identity. 

The standard deviation of this added noise is selected so as to avoid a significant drop in the original model's accuracy. Let us note that even though the selected standard deviation might depend on the victim network, several CNNs approximating the identity are trained independently from the victim network, and the victim can then select one or several CNNs adapted to the network at hand.  

Since the noise added is fixed, it only constitutes a translation of the victim hyperplanes, and can be approximated by the CNN through an increase in the bias $\beta$. We avoid this case by bounding the bias to a small value ($||\beta||_2 < \epsilon$) or eliminating the bias ($\beta = 0$). This makes the learning task more complex, and forces the filter values themselves to change, thus preventing the introduced hyperplanes from being simple translations of the original ones.

Let us consider, for instance, the case where one convolutional layer is introduced. As before, let $C$ be the matrix associated to the layer and $\mathcal{N}$ be the fixed Gaussian noise. Then the optimization problem becomes:

$$
\sum_{1 \leq k \leq m} C_{i, k} x_k = x_i + \mathcal{N}_i \text{  } \forall 1 \leq i \leq n 
$$
where $n$ is the number of output neurons and $m$ is the number of input neurons. The only element that is independent of the input is the noise $\mathcal{N}$. This makes this system of equations impossible to solve for all inputs $x$. Thus, the solution $C^*$ provided by the CNN is such that:
$$
\sum_{1 \leq k \leq m} C_{i, k} x_k = x_i + \mathcal{N^*}_i(x) \text{  } \forall 1 \leq i \leq n 
$$
where $\mathcal{N^*}$ is a noise close to $\mathcal{N}$ but depends on the input. $C^*$ leads to hyperplanes for the various inputs which cannot be translations of the input hyperplanes. This implies that the newly introduced hyperplanes intersect the original ones, increasing the chances of modifying the polytopes formed by all the model's boundaries. This explanation generalizes to the case of several layers. Indeed, in the general case, the optimization problem for $k$ convolutional layers without a bias becomes:
$$
f(x) = x_i + \mathcal{N^*}_i(x) \text{  } \forall 1 \leq i \leq n
$$
with $f(x) = ReLU(C_k(ReLU(...ReLU(C_1 x))))$, where $C_j$ is the matrix associated to the $j-th$ layer.

In order to further prevent the introduced hyperplanes from being too far from the working space, we add Batch Normalization layers after each convolutional layer.

To ensure the hyperplanes have indeed changed, we measure the influence on adversarial examples. Adversarial attackers find the shortest path from one prediction class to another. This path depends on the subdivsion of the space by the original model's hyperplanes. Thus, changes in the said subdivision leads to different adversarial samples. 
Conversely, if two models lead to the same subdivision of the space, then adversarial examples remain the same for both models.
Therefore, in Sec.~\ref{experiments}, we measure the impact of the added CNN on both the original model's accuracy and the adversarial samples. Let us note that adding the CNN to the model we want to protect does not prevent adversarial examples in itself: it only changes some of them.

\section{Experiments} \label{experiments}

In this section, we detail the model we want to protect as well as the added CNN. Then, we measure the impact of the added layers on the model to protect by counting the number of adversarial samples which do not generalize to the protected model. 

\subsection{Description of the NN models used}
For our CNN approximating a noisy identity -- called parisitic CNN from now on, we consider a CNN with 4 convolutional layers, $5 \times 5$ filter sizes and separated by ReLU activation functions (see Fig. \ref{CNN_no_batch}). In a second model, we separate the convolutional layers from their activation by Batch Normalization layers (see Fig. \ref{CNN_batch}). 
The batch normalizations in this second model normalize their input, ensuring a mean of 0 and a standard deviation of 1. This increases the chances of the ReLU functions being activated. The first three convolutional layers have two channels, while the last one only has one. We train this model over 10,000 random inputs $\{x_i \in [0, 1]^n\}_{1 \leq i \leq 10,000} $ of size $n = 16 \times 16$. In our experiments, we select the $n$ input neurons as the first or the last ones from the previous layer, but they can be selected at random among the previous layer's neurons. For a given training, we fix $\mathcal{N}$ a Gaussian noise, and we set the labels to be $\{x_i + \mathcal{N}\}_{1 \leq i \leq 10,000}$.

\begin{figure}[ht]
\centering
\includegraphics[scale=0.375]{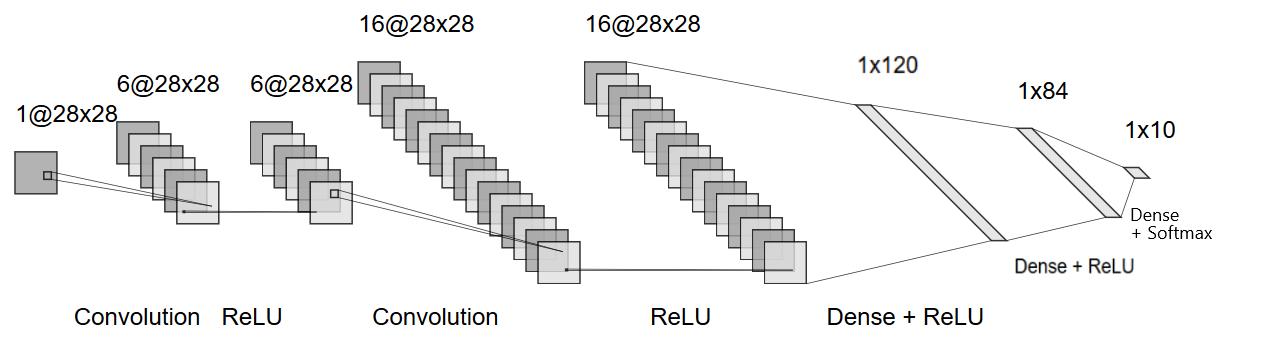}
\caption{LeNet architecture, as in \cite{lenet}.}
\label{Lenet_no_batch}
\end{figure}

\begin{figure}[ht]
\centering
\includegraphics[scale=0.4]{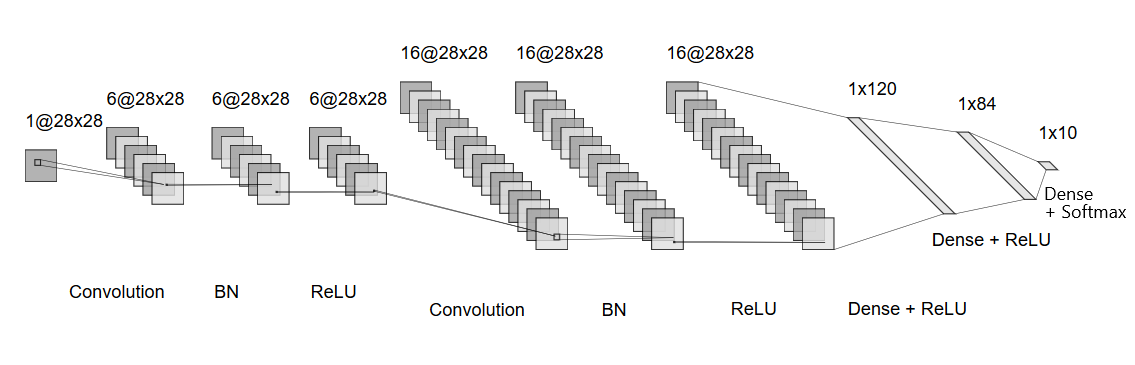}
\caption{LeNet architecture, as in \cite{lenet}, where a batch normalization (BN) layer is added after each convolution.}
\label{Lenet_batch}
\end{figure}

The model to protect is a LeNet architecture \cite{lenet} trained on the MNIST dataset \cite{mnist} (see Fig. \ref{Lenet_no_batch}). We also consider a second model where we introduce batch normalization layers after the convolutional layers of the LeNet arcitecture (see Fig. \ref{Lenet_batch}). We denote $VM$ the victim LeNet architecture and $VM_{batch}$ the architecture where batch normalization layers have been added. 

$VM$ has an accuracy of 98.78\% on the MNIST dataset, while $VM_{batch}$'s accuracy is of 99.11\%. 

\subsection{Adversarial examples} \label{adv_ex}
Several methods enable an attacker to compute adversarial samples \cite{goodfellow,pgd,deepfool,vat}. In this paper, we use the Fast Gradient Sign Method introduced by Goodfellow et al. \cite{goodfellow} to determine adversarial samples for our LeNet architecture. Given an input $x$ in the MNIST dataset, the algorithm computes the adversarial example $x_{adv}$ as follows:
\begin{equation}\label{fgsm}
x_{adv} = x + \epsilon \times sign(\nabla_x \mathcal{L}(\theta, x, y))
\end{equation}

where $\mathcal{L}$ is the victim model's loss function, $\theta$ is its vector of parameters and $y$ is $x$'s true prediction. 

Since adversarial examples are based on the subdivision of the space by the neurons' hyperplanes, a modification of those examples is a good indicator that the said subdivision has indeed been changed by the added CNN. Thus, we compute adversarial samples for the first 200 images of the MNIST set using the FGSM method and measure the percentage of examples which do not generalize to the modified model. For the FGSM method, we start with $\epsilon = 0.05$ and increase it by $0.05$ until the computed $x_{adv}$ is indeed an adversarial example for the original model. 

Furthermore, let us denote $M_{adv}$ the percentage of adversarial examples for the original model which are no longer adversarial for the protected model. 

\subsection{Results}

We test the two original models considered with the added parasitic CNNs, without a bias $\beta$ or with the constraint that $||\beta||_2 < 0.05$. In Table \ref{tab}, the parasitic CNN is added to the first $16 \times 16$ neurons of the second convolutional layer. The parasitic CNNs approximate the identity to which a centered Gaussian noise with standard deviation $\sigma = 0.2$ has been added. 
In every case, we observe a change in the adversarial examples. Let us note that we only count the number of adversarial samples for the original model that are no longer adversarial for the protected CNN. There are also examples which are adversarial for both models, but with different predictions. 

In all cases, $M_{adv}$ -- as defined in Sec.~\ref{adv_ex} --, is higher or equal to  $12 \%$, and the accuracy of the protected model is very close to the original one. This shows that the boundaries between classes -- which are the result of the various layers' hyperplanes -- have changed. The summary of our results can be found in Table~\ref{tab}.

\begin{table}[ht]
\caption{Measurement of the accuracy, and percentage of the adversarial samples that are no longer adversarial for the protected CNN ($M_{adv}$). All tests are made on the MNIST dataset \cite{mnist}, and the parasitic CNNs approximate the identity to which a centered Gaussian noise with a standard deviation of 0.2 was added. BN denotes Batch normalization. All parasitic CNNs were added after the second convolutional layer of the original model. }
\begin{tabular}{m{2cm} m{1.4cm} m{1.4cm} m{1.8cm} m{2.2cm} m{1.2cm} m{1.2cm} }
\hlineB{2}
\textbf{CNN Location} & \textbf{Original Model} & \textbf{Original Accuracy} & \textbf{Identity CNN} &  \textbf{Bias constraints} & \textbf{New accuracy} & \boldmath{$M_{adv}$}  \\
\hlineB{2}
\multirow{8}{1.5cm}{After BN and activation (if BN)} & \multirow{3}{1.2cm}{$VM$ (Fig. \ref{Lenet_no_batch})} & \multirow{3}{*}{98.78\%} & \multirow{2}{*}{Without BN} & $||\beta||_2 < 0.05$ & 98.69\% & 24.5\%  \\
	&	& & & No bias & 98.7 \% & 19 \% \\
\cline{4-7}
   	  &	& 		& \multirow{2}{*}{With BN} & $||\beta||_2 < 0.05$ & 98.50 \% & 28\% \\
	  &			&  & & No bias & 98.67\% & 22\% \\
\cline{2-7}
& \multirow{4}{1.2cm}{$VM_{batch}$ (Fig. \ref{Lenet_batch})} 	& \multirow{4}{*}{99.11\%} & \multirow{2}{*}{Without BN} & $||\beta||_2 < 0.05$ & 99.24\% & 17.5\% \\									
						& & & & No bias & 99.14\% & 14\% \\
						\cline{4-7}
						& & & \multirow{2}{*}{With BN} &  $||\beta||_2 < 0.05$ & 99.18\% & 17\% \\
						& & & & No bias & 99.15 \% & 12 \% \\
						\cline{4-7}
\hline
\multirow{4}{1.5cm}{Before BN and activation (if BN)} & \multirow{4}{1.2cm}{$VM_{batch}$ (Fig. \ref{Lenet_batch})} 	& \multirow{4}{*}{99.11\%} & \multirow{2}{*}{Without BN} & $||\beta||_2 < 0.05$ & 96.64\% & 37.5\% \\	
						& & & & No bias & 98.13\% & 39\% \\
						\cline{4-7}
						& & & \multirow{2}{*}{With BN} &  $||\beta||_2 < 0.05$ & 99.05\% & 27.5\% \\
						& & & & No bias & 99.16 \% & 14 \% \\
\hlineB{2}
\end{tabular}
\label{tab}
\end{table}

As Table~\ref{tab} shows it, inserting a CNN trained to learn a Gaussian noise added to the identity can lead to a modification of the polytopes formed by the original model's hyperplanes, with only a slight drop in the accuracy. 

It is interesting to note that the parasitic CNNs trained with no bias, although they incur a lower $M_{adv}$, entail either a lower drop in the accuracy than the CNNs learnt with a small bias, or an increased accuracy. This might 
 be explained by the fact that the CNN with no bias cannot learn a noise independent of the input, and will therefore tend to get closer to the non-noisy identity. Furthermore, the ability for the parasitic CNN to operate a translation thanks to the small bias can explain the small drop in the accuracy that we observe. However, despite this added possibility, the CNN with a small bias still changes the slope of the hyperplanes, as the drop in the accuracy is not steep enough to justify the high $M_{adv}$.

It is also possible to add several parasitic CNNs to a given victim NN. This might result in a higher protection, with no -- or a small -- drop in the accuracy. Since the parasitic CNNs are already trained, the cost of adding these CNNs remains small, and is equal to the additional computations required for inference. On $VM_{batch}$, we try adding two parasitic CNNs after the same layer, one parasitic CNN after the first and the second layers, as well as two parasitic layers after one layer and a third CNN after a second layer. Table \ref{several} gives an example of accuracy and $M_{adv}$ obtained in various cases where the parasitic CNNs are added after the second convolutional layer from $VM_{batch}$, either before or after the batch normalization layer and the activation function. Let us note that once again, the standard deviation of the added noise is 0.2 in all cases. Moreover, when there are two parasitic CNNs at the same location, the first is applied to the first neurons and the second is applied to the last neurons of the victim layer. 

\begin{table}[ht]
\caption{Measurement of the accuracy, and percentage of the adversarial samples that are no longer adversarial for the protected NN ($M_{adv}$). Several parasitic CNNs were added to the victim NN. All tests are made on the MNIST dataset \cite{mnist}, and the parasitic CNNs approximate the identity to which a centered Gaussian noise with a standard deviation of 0.2 was added. BN denotes Batch normalization. The parasitic CNNs are added after the second convolutional layer of $VM_{batch}$. We add them before, after, or both before and after the BN layer and activation function. The original accuracy for $VM_{batch}$ is 99.11\%. $Small$ means that the constraint on the bias $\beta$ is $||\beta||_2 < 0.05$.}
\begin{tabular}{ m{2.3 cm}  m{1.6cm} m{1.6cm} | m{1.6cm} m{1.6cm} | m{1.2cm} m{1.2cm}}
\hlineB{2}
\multicolumn{5}{c}{Parasitic CNNs' Locations} & \multicolumn{2}{c}{Accuracy and $M_{adv}$} \\
\hlineB{2}
& \multicolumn{2}{>{\centering}p{.27\textwidth}}{\textbf{$2^{nd}$ layer, \newline Before BN  \newline and activation}} & \multicolumn{2}{>{\centering}p{.27\textwidth}}{\textbf{$2^{nd}$ layer, \newline After BN \newline and activation}} &  \textbf{New \newline accuracy} & \boldmath{ $M_{adv}$}  \\
\hlineB{2}
 & With BN? & With Bias? & With BN? & With Bias? & \multicolumn{2}{c}{} \\ 
\hlineB{2}
First $n$ neurons & BN & Small & No BN & Small & 99\% & 31\%  \\
\hline
First $n$ neurons & BN & Small & BN & Small	& 98.98\%& 37\% \\
\hline
First $n$ neurons & BN & Small & No BN & No bias & 98.93\% & 31.5\% \\
\hline
First $n$ neurons & BN & Small & BN & No bias	& 98.99\% & 31\% \\
\hline
First $n$ neurons & BN & No bias & BN & No bias & 98.96\% & 28.5\% \\
\hlineB{2}
First $n$ neurons & BN & Small & \multirow{2}{*}{-} & \multirow{2}{*}{-} & \multirow{2}{*}{99.05\%} & \multirow{2}{*}{31\%} \\ 
Last $n$ neurons & BN & No bias &  &  & & \\	
\hline								
First $n$ neurons & \multirow{2}{*}{-} & \multirow{2}{*}{-} & BN & Small & \multirow{2}{*}{99.17\%} & \multirow{2}{*}{27.5\%} \\
Last $n$ neurons & & &  BN & No bias & & \\
\hline
First $n$ neurons & \multirow{2}{*}{-} & \multirow{2}{*}{-} & BN & No bias & \multirow{2}{*}{99.15\%} & \multirow{2}{*}{27\%} \\
Last $n$ neurons & & & BN & No bias & & \\
\hline
First $n$ neurons & \multirow{2}{*}{-} & \multirow{2}{*}{-} & No BN & No bias & \multirow{2}{*}{99.19\%} & \multirow{2}{*}{25.5\%} \\
First $n$ neurons & & & No BN & Small & & \\
\hline
First $n$ neurons & BN & Small  & BN & Small & \multirow{2}{*}{98.94\%} & \multirow{2}{*}{40\%} \\
Last $n$ neurons & BN & No bias & & \\
\hline
First $n$ neurons & BN & Small & BN & Small & \multirow{2}{*}{99.03\%} & \multirow{2}{*}{38.5\%} \\
Last $n$ neurons & - & - & BN & Small & & \\
\hline
First $n$ neurons & BN & Small  & BN & Small & \multirow{2}{*}{98.89\%} & \multirow{2}{*}{43\%} \\
Last $n$ neurons & BN & No bias & - & - & &  \\
\hlineB{2}
\end{tabular}
\label{several}
\end{table}

We observe that adding a parasitic CNN to the first victim layer did not improve much the results, as there was almost no impact be it on the accuracy or $M_{adv}$.

\section{Conclusion} \label{conclusion}

In this paper, we introduce a simple but effective countermeasure to thwart the recent wave of attacks \cite{DBLP:journals/corr/abs-2003-04884,DBLP:conf/fat/MilliSDH19,DBLP:journals/corr/abs-1910-00744,DBLP:journals/corr/abs-1909-01838} aiming at the extraction of NN models through an oracle access.

As a line of further research, we want to investigate the gain we get by mounting these attacks over quantized NNs \cite{DBLP:journals/jmlr/HubaraCSEB17,DBLP:journals/corr/HanMD15,DBLP:journals/corr/GongLYB14,DBLP:journals/corr/ZhouNZWWZ16,DBLP:conf/cvpr/JacobKCZTHAK18}. Indeed, in the non-quantized case, a great care should be taken dealing with floating point imprecision with real numbers machine representation, as   reported, for instance, by \cite{DBLP:journals/corr/abs-2003-04884}. Today, Quantized NNs share almost the same accuracy as the floating-point ones.  By doing that, we are coming a step closer to differential cryptanalysis \cite{DBLP:books/daglib/0032320} performed against symmetric ciphers and which serves as an inspiration of \cite{DBLP:journals/corr/abs-2003-04884}. While our protection will still be relevant, we want to explore more cryptographic techniques as alternatives.

\bibliographystyle{splncs04}
\bibliography{biblio_vc}

\begin{thebibliography}{10}
\providecommand{\url}[1]{\texttt{#1}}
\providecommand{\urlprefix}{URL }
\providecommand{\doi}[1]{https://doi.org/#1}

\bibitem{adv_survey}
Akhtar, N., Mian, A.S.: Threat of adversarial attacks on deep learning in
  computer vision: {A} survey. {IEEE} Access  \textbf{6},  14410--14430 (2018).
  \doi{10.1109/ACCESS.2018.2807385},
  \url{https://doi.org/10.1109/ACCESS.2018.2807385}

\bibitem{DBLP:conf/uss/BatinaBJP19}
Batina, L., Bhasin, S., Jap, D., Picek, S.: {CSI} {NN:} reverse engineering of
  neural network architectures through electromagnetic side channel. In:
  {USENIX} Security Symposium. pp. 515--532. {USENIX} Association (2019)

\bibitem{DBLP:books/daglib/0032320}
Biham, E., Shamir, A.: Differential Cryptanalysis of the Data Encryption
  Standard. Springer (1993)

\bibitem{DBLP:journals/corr/abs-2002-11021}
Breier, J., Jap, D., Hou, X., Bhasin, S., Liu, Y.: {SNIFF:} reverse engineering
  of neural networks with fault attacks. CoRR  \textbf{abs/2002.11021} (2020)

\bibitem{DBLP:journals/corr/abs-2003-04884}
Carlini, N., Jagielski, M., Mironov, I.: Cryptanalytic extraction of neural
  network models. CoRR  \textbf{abs/2003.04884} (2020)

\bibitem{DBLP:journals/corr/GongLYB14}
Gong, Y., Liu, L., Yang, M., Bourdev, L.D.: Compressing deep convolutional
  networks using vector quantization. CoRR  \textbf{abs/1412.6115} (2014)

\bibitem{goodfellow}
Goodfellow, I.J., Shlens, J., Szegedy, C.: Explaining and harnessing
  adversarial examples. In: Bengio, Y., LeCun, Y. (eds.) 3rd International
  Conference on Learning Representations, {ICLR} 2015, San Diego, CA, USA, May
  7-9, 2015, Conference Track Proceedings (2015),
  \url{http://arxiv.org/abs/1412.6572}

\bibitem{DBLP:journals/corr/HanMD15}
Han, S., Mao, H., Dally, W.J.: Deep compression: Compressing deep neural
  network with pruning, trained quantization and huffman coding. In: {ICLR}
  (2016)

\bibitem{pni}
He, Z., Rakin, A.S., Fan, D.: Parametric noise injection: Trainable randomness
  to improve deep neural network robustness against adversarial attack. In:
  {IEEE} Conference on Computer Vision and Pattern Recognition, {CVPR} 2019,
  Long Beach, CA, USA, June 16-20, 2019. pp. 588--597. Computer Vision
  Foundation / {IEEE} (2019). \doi{10.1109/CVPR.2019.00068},
  \url{http://openaccess.thecvf.com/content\_CVPR\_2019/html/He\_Parametric\_Noise\_Injection\_Trainable\_Randomness\_to\_Improve\_Deep\_Neural\_Network\_CVPR\_2019\_paper.html}

\bibitem{DBLP:journals/corr/abs-2002-06776}
Hong, S., Davinroy, M., Kaya, Y., Dachman{-}Soled, D., Dumitras, T.: How to 0wn
  {NAS} in your spare time. CoRR  \textbf{abs/2002.06776} (2020)

\bibitem{DBLP:journals/jmlr/HubaraCSEB17}
Hubara, I., Courbariaux, M., Soudry, D., El{-}Yaniv, R., Bengio, Y.: Quantized
  neural networks: Training neural networks with low precision weights and
  activations. J. Mach. Learn. Res.  \textbf{18},  187:1--187:30 (2017)

\bibitem{DBLP:conf/cvpr/JacobKCZTHAK18}
Jacob, B., Kligys, S., Chen, B., Zhu, M., Tang, M., Howard, A.G., Adam, H.,
  Kalenichenko, D.: Quantization and training of neural networks for efficient
  integer-arithmetic-only inference. In: {CVPR}. pp. 2704--2713. {IEEE}
  Computer Society (2018)

\bibitem{DBLP:journals/corr/abs-1909-01838}
Jagielski, M., Carlini, N., Berthelot, D., Kurakin, A., Papernot, N.:
  High-fidelity extraction of neural network models. CoRR
  \textbf{abs/1909.01838} (2019)

\bibitem{malware}
Kaspersky: Machine learning methods for malware detection. {Whitepaper}
  (2020),
  \url{https://media.kaspersky.com/en/enterprise-security/Kaspersky-Lab-Whitepaper-Machine-Learning.pdf}

\bibitem{lenet}
Lecun, Y., Bottou, L., Bengio, Y., Haffner, P.: Gradient-based learning applied
  to document recognition. In: Proceedings of the IEEE. pp. 2278--2324 (1998)

\bibitem{mnist}
LeCun, Y., Cortes, C., Burges, C.: Mnist handwritten digit database. ATT Labs
  [Online]. Available: http://yann. lecun. com/exdb/mnist  \textbf{2} (2010)

\bibitem{pgd}
Madry, A., Makelov, A., Schmidt, L., Tsipras, D., Vladu, A.: Towards deep
  learning models resistant to adversarial attacks. In: 6th International
  Conference on Learning Representations, {ICLR} 2018, Vancouver, BC, Canada,
  April 30 - May 3, 2018, Conference Track Proceedings. OpenReview.net (2018),
  \url{https://openreview.net/forum?id=rJzIBfZAb}

\bibitem{DBLP:conf/fat/MilliSDH19}
Milli, S., Schmidt, L., Dragan, A.D., Hardt, M.: Model reconstruction from
  model explanations. In: {FAT}. pp.~1--9. {ACM} (2019)

\bibitem{vat}
Miyato, T., Maeda, S., Koyama, M., Ishii, S.: Virtual adversarial training: {A}
  regularization method for supervised and semi-supervised learning. {IEEE}
  Trans. Pattern Anal. Mach. Intell.  \textbf{41}(8),  1979--1993 (2019).
  \doi{10.1109/TPAMI.2018.2858821},
  \url{https://doi.org/10.1109/TPAMI.2018.2858821}

\bibitem{deepfool}
Moosavi{-}Dezfooli, S., Fawzi, A., Frossard, P.: Deepfool: a simple and
  accurate method to fool deep neural networks. CoRR  \textbf{abs/1511.04599}
  (2015), \url{http://arxiv.org/abs/1511.04599}

\bibitem{adv}
Papernot, N., McDaniel, P.D., Jha, S., Fredrikson, M., Celik, Z.B., Swami, A.:
  The limitations of deep learning in adversarial settings. In: {IEEE} European
  Symposium on Security and Privacy, EuroS{\&}P 2016, Saarbr{\"{u}}cken,
  Germany, March 21-24, 2016. pp. 372--387. {IEEE} (2016).
  \doi{10.1109/EuroSP.2016.36}, \url{https://doi.org/10.1109/EuroSP.2016.36}

\bibitem{DBLP:journals/corr/abs-1910-00744}
Rolnick, D., K{\"{o}}rding, K.P.: Reverse-engineering deep relu networks. CoRR
  \textbf{abs/1910.00744} (2019)

\bibitem{DBLP:journals/corr/abs-1901-10861}
Shamir, A., Safran, I., Ronen, E., Dunkelman, O.: A simple explanation for the
  existence of adversarial examples with small hamming distance. CoRR
  \textbf{abs/1901.10861} (2019)

\bibitem{vgg}
Simonyan, K., Zisserman, A.: Very deep convolutional networks for large-scale
  image recognition. CoRR  (2015), \url{http://arxiv.org/abs/1409.1556}

\bibitem{DBLP:journals/corr/abs-1808-04761}
Yan, M., Fletcher, C.W., Torrellas, J.: Cache telepathy: Leveraging shared
  resource attacks to learn {DNN} architectures. CoRR  \textbf{abs/1808.04761}
  (2018)

\bibitem{DBLP:journals/corr/abs-1902-04698}
Zhang, C., Bengio, S., Hardt, M., Singer, Y.: Identity crisis: Memorization and
  generalization under extreme overparameterization. CoRR
  \textbf{abs/1902.04698} (2019)

\bibitem{DBLP:journals/corr/ZhouNZWWZ16}
Zhou, S., Ni, Z., Zhou, X., Wen, H., Wu, Y., Zou, Y.: Dorefa-net: Training low
  bitwidth convolutional neural networks with low bitwidth gradients. CoRR
  \textbf{abs/1606.06160} (2016)

\end{thebibliography}

\end{document}